\documentclass[journal]{IEEEtran}

\usepackage{amsmath,amsfonts}
\usepackage{algorithmic}
\usepackage{algorithm}
\usepackage{array}
\usepackage[caption=false,font=normalsize,labelfont=sf,textfont=sf]{subfig}
\usepackage{textcomp}
\usepackage{enumerate}
\usepackage{stfloats}
\usepackage{url}
\usepackage{verbatim}
\usepackage{graphicx}
\usepackage{enumitem}
\usepackage{cite}
\usepackage{hyperref}
\usepackage{multirow}
\usepackage[table,xcdraw]{xcolor}
\usepackage{lscape}
\usepackage{colortbl}
\usepackage{booktabs}
\usepackage{adjustbox}
\usepackage{array}
\usepackage{listings}
\usepackage{xcolor}
\usepackage{enumitem}
\usepackage{tabularx}

\begin{document}

\title{Impact of Comprehensive Data Preprocessing on Predictive Modelling of COVID-19 Mortality}

\author{
    \begin{minipage}[t]{0.45\textwidth}
        \centering
        \textbf{Sangita Das} \\
        Syamaprasad College \\
        dassangita844@gmail.com
    \end{minipage}
    \begin{minipage}[t]{0.45\textwidth}
        \centering
        \textbf{Subhrajyoti Maji} \\
        Independent Researcher \\
        subhrajyoti.maji@gmail.com
    \end{minipage}
}

\maketitle


\begin{abstract}
\label{section: Abstract}
Accurate predictive models are crucial for analysing COVID-19 mortality trends. This study evaluates the impact of a custom data preprocessing pipeline on ten machine learning models predicting COVID-19 mortality using data from Our World in Data (OWID). Our pipeline differs from a standard preprocessing pipeline through four key steps. Firstly, it transforms weekly reported totals into daily updates, correcting reporting biases and providing more accurate estimates. Secondly, it uses localised outlier detection and processing to preserve data variance and enhance accuracy. Thirdly, it utilises computational dependencies among columns to ensure data consistency. Finally, it incorporates an iterative feature selection process to optimise the feature set and improve model performance. Results show a significant improvement with the custom pipeline: the \emph{MLP Regressor} achieved a test RMSE of 66.556 and a test R² of 0.991, surpassing the \emph{DecisionTree Regressor} from the standard pipeline, which had a test RMSE of 222.858 and a test R² of 0.817. These findings highlight the importance of tailored preprocessing techniques in enhancing predictive modelling accuracy for COVID-19 mortality. Although specific to this study, these methodologies offer valuable insights into diverse datasets and domains, improving predictive performance across various contexts.

\end{abstract}

\begin{IEEEkeywords}
COVID-19 Mortality Prediction, Data Preprocessing, Custom Pipeline, Feature Selection, Predictive Modeling, Machine Learning.
\end{IEEEkeywords}
\section{Introduction}
\label{section: Introduction}

The COVID-19 pandemic has profoundly disrupted global health systems and daily life, presenting unprecedented challenges in disease management and resource allocation. Since the emergence of Severe Acute Respiratory Syndrome Coronavirus 2 (SARS-CoV-2) in December 2019, the pandemic has resulted in extensive morbidity and mortality worldwide. In India, where the first cases were reported in January 2020 \cite{andrews2020first}, the impact has been particularly severe, with over 45 million confirmed cases and more than 530,000 deaths as of July 2024 \cite{owidcoronavirus}. The pandemic disproportionately affected older adults and individuals with pre-existing health conditions such as hypertension, diabetes, and cardiovascular disease \cite{mueller2020does, petretto2020ageing}.

As the pandemic progressed, the need for accurate predictive models to guide healthcare planning and resource allocation became increasingly critical \cite{an2020machine, rustam2020covid,das2020predicting, booth2021development, subudhi2021comparing, mahdavi2021machine, krithika2021comparative}. The foundation of these models depends heavily on the quality of data and the preprocessing steps applied, both of which are crucial for ensuring the reliability of predictions \cite{ wang2019progress, rahm2000data, guyon2003introduction,  bergstra2011algorithms}. Robust preprocessing transforms raw data into meaningful features through techniques such as dropping irrelevant columns \cite{winston2022exploring, moustakidis2022identifying, sha2023assessing, eryilmazprediction, pourhomayoun2021predicting}, creating new features \cite{schrarstzhaupt2024interactive}, and renaming columns \cite{beattie2021worldwide}. Imputing missing values using
zero, mean, or median values, along with interpolation and extrapolation methods \cite{hu2020early, khan2021computational, zubair2021efficient, xia2023analysis, jerez2010missing, batista2003analysis, vaid2020machine, moustakidis2022identifying, chowdhury2021early, pan2020prognostic} ensures that gaps in the data do not compromise the model accuracy. Outlier processing—whether through dropping, trimming, transforming, or winsorizing  \cite{turlapati2020outlier, brzezinska2021outliers, herawati2022implementation, xia2023analysis}, preserves data integrity by addressing anomalies. Feature selection, using methods such as correlation \cite{moulaei2022comparing}, principal component analysis (PCA) \cite{pan2020prognostic}, random feature exploration \cite{moustakidis2022identifying}, ranking features through univariate and multivariate filter methods, and wrapper methods \cite{pourhomayoun2021predicting, tulu2023machine} optimizes model performance by reducing multicollinearity and eliminating redundant features \cite{gambhir2020regression}. These steps are essential for enhancing the reliability and accuracy of predictive models \cite{garcia2016big, gudivada2017data}.

Despite significant advancements in predictive modelling, many current approaches prioritize model development over thorough data preprocessing. This oversight often results in a superficial understanding of the data, neglecting its inherent patterns, relationships, and dependencies, crucial for developing robust models. For instance, failing to account for weekly reporting patterns can distort model training and degrade performance. Ignoring computational dependencies between columns can introduce inconsistencies, undermining the integrity of the entire modelling process. Moreover, using global outlier detection methods with fixed thresholds, such as z-scores, fails to accommodate the local variability inherent in time-series data like COVID-19, resulting in inaccurate anomaly detection. Finally, neglecting rigorous feature selection can lead to multicollinearity, and feature redundancy, and result in issues such as underfitting or overfitting, all of which severely model accuracy and reliability.

\begin{figure*}[!t]
\centering
\captionsetup[subfloat]{labelfont=scriptsize,textfont=scriptsize}
\subfloat[Standard Preprocessing Pipeline \label{subfigure: standard pipeline}]{\includegraphics[width=0.292\linewidth, trim={0 1.2cm 0 1.2cm}]{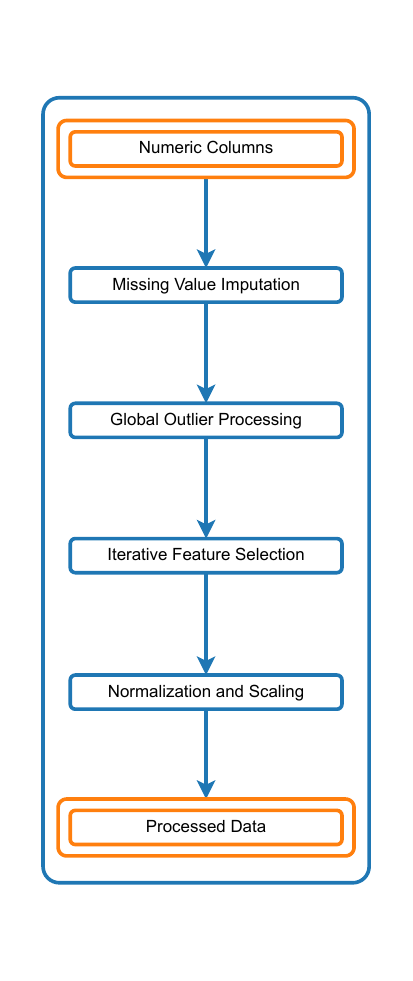}}
\subfloat[Custom Preprocessing Pipeline \label{subfigure: custom pipeline}]{\includegraphics[width=0.69\linewidth]{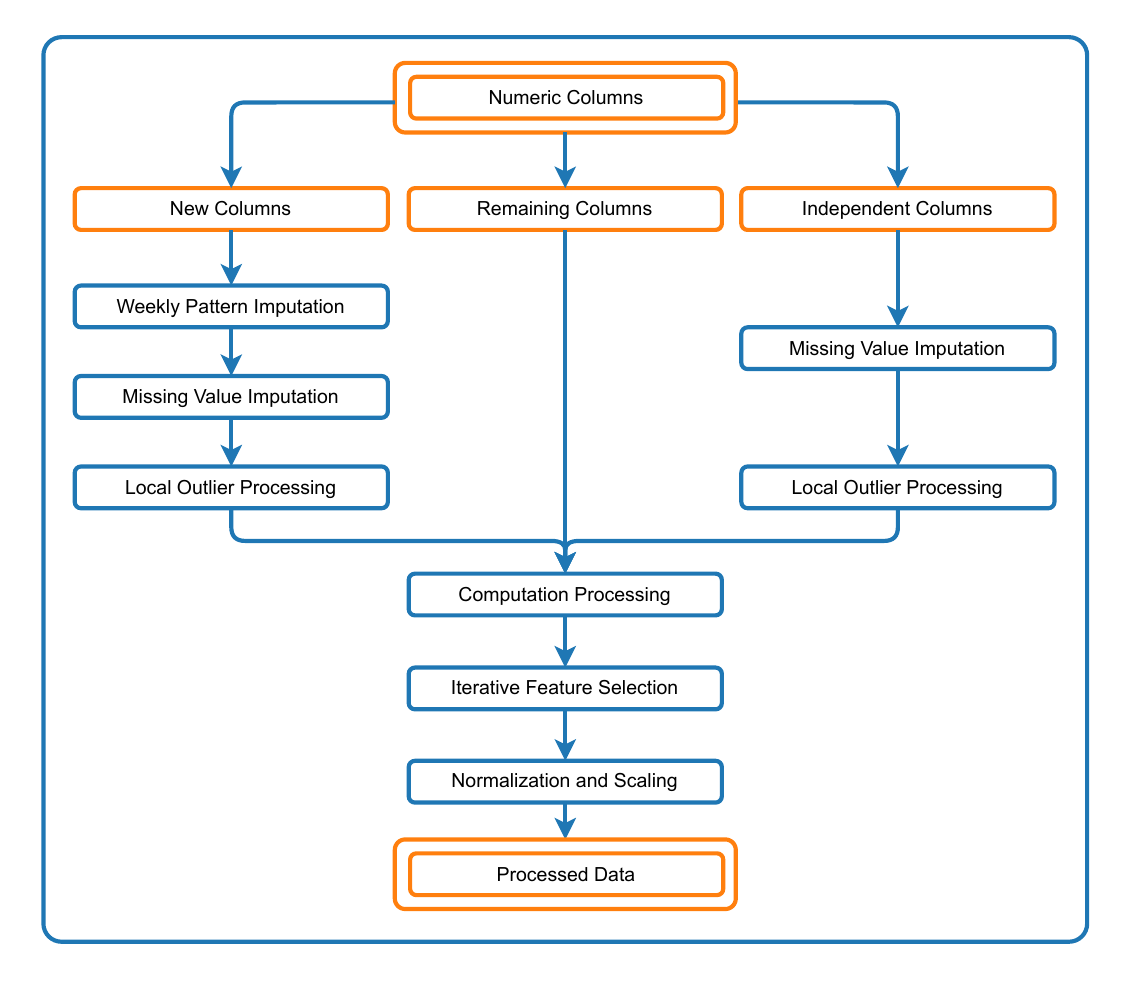}}%
\caption{COVID Data Preprocessing Pipelines for (a) Standard and (b) Custom approaches}
\label{figure: preprocessing pipelines}
\end{figure*}

This study addresses these gaps by introducing a novel custom preprocessing pipeline (see Figure \ref{subfigure: custom pipeline}) designed to address the unique challenges of COVID-19 datasets. The primary objective is to enhance data consistency and coherence through comprehensive exploration, meticulous cleaning, and effective preprocessing. The custom pipeline incorporates several innovative steps:

\paragraph{Weekly Pattern Imputation} Transforms weekly reported totals into daily updates by averaging the weekly sum and distributing it evenly across each day, thereby correcting summing errors and addressing reporting inconsistencies.

\paragraph{Local Outlier Processing} Uses a rolling window approach for local outlier detection, which accommodates local data variability rather than relying on standard global outlier detection methods.

\paragraph{Computation Processing} Calculates column values by leveraging inter-column dependencies, thereby preserving data consistency.

\paragraph{Iterative Feature Selection}
Employs advanced techniques such as Permutation Feature Importance (PFI) \cite{michelucci2024feature}, Mutual Information (MI) \cite{kraskov2004estimating}, Single Feature Impact (SFI) \cite{guyon2003introduction}, and the Variance Inflation Factor (VIF) \cite{greene2003econometric} to systematically eliminate redundant and collinear features while retaining the most significant ones. This approach refines the feature set by balancing predictive power, addressing multicollinearity, and improving overall model performance.

Our results show that all tested models benefit from the custom preprocessing pipeline, with non-linear models achieving the highest scores as anticipated. The consistent and stable performance across training, validation, and test sets highlights the novel contributions of this custom preprocessing pipeline in improving predictive accuracy and reliability. These findings offer valuable insights for researchers, policymakers, and healthcare professionals, supporting improved decision-making in pandemic management and future health crises.

The remainder of this paper is organized as follows: Section \ref{section: Data and Methodology} outlines the data sources, preprocessing techniques, and machine learning models and evaluation strategies used in this study. Section \ref{section: Results And Discussion} compares the model performance across different preprocessing pipelines and discusses the impact of various preprocessing techniques. Finally, Section \ref{section: Conclusion} summarizes the findings, highlights the contributions, and offers recommendations for future work.
\section{Data and Methodology}
\label{section: Data and Methodology}

\subsection{Data}
The dataset used in this study is sourced from \href{https://ourworldindata.org/coronavirus}{Our World In Data (OWID)} \cite{owidcoronavirus}, and comprises over 400,000 rows and 67 columns, covering data from around the world. The dataset includes information from 244 countries, 6 continents, the European Union, and various income levels. Our analysis focuses specifically on India, using data collected from January 5, 2020, to August 11, 2024, amounting to 1,680 records.

The dataset includes one date column, four categorical columns (e.g., $iso\_code$, $location$, $continent$, and $tests\_units$), and 62 numeric columns that provide a detailed view of COVID-19 trends and influencing factors. Among the numeric columns, 12 are empty, 15 are constant, and 35 exhibit variability, which provides a rich source for analysis.

\subsection{Standard Preprocessing Pipeline}

In the \emph{standard preprocessing pipeline} (see Figure \ref{subfigure: standard pipeline}), we applied traditional methods for dataset preparation. The process included the following steps:

\subsubsection{Extracting Numeric Features}
We focused on numeric columns as categorical columns remained constant across countries and did not contribute to the analysis.

\subsubsection{Missing Value Imputation}
We handled missing values using a two-step approach. Firstly, we applied linear interpolation and extrapolation to estimate the missing values. Then, we filled any remaining gaps with zero imputation. The orange line in Figure \ref{figure: computation processing of positive_rate} visually illustrates how we applied this process to impute $positive\_rate$.

\subsubsection{Global Outlier Processing}
We identified outliers using a z-score threshold of 2 and then processed them by applying linear interpolation. Figure \ref{subfigure: global outlier processing} shows how we used this method for $new\_vaccinations$.

\subsubsection{Iterative Feature Selection}
We systematically identified the most relevant features for our predictive model and addressed multicollinearity through iterative feature selection, as detailed in Algorithm \ref{algorithm: iterative_feature_selection}. The key steps included:

\paragraph{Correlation Filtering}
We removed features with correlations above 0.8 ($corr_{th} = 0.8$) to minimise redundancy and multicollinearity. We also discarded constant features (zero correlations) and empty features (undefined correlations), because they do not provide useful information.

\paragraph{Calculating Importance Metrics}
We assessed the importance of each remaining feature using three metrics:
\begin{itemize}
    \item \textbf{Permutation Feature Importance (PFI):} Measures the impact on model performance when feature values are shuffled, reflecting overall predictive power.
    \item \textbf{Mutual Information (MI):} Quantifies the information a feature provides about the target, capturing both linear and non-linear dependencies.
    \item \textbf{Single Feature Impact (SFI):} Assesses the predictive power of each feature in isolation, offering a baseline view of its importance.
\end{itemize}

These metrics provided a comprehensive view of feature importance. PFI captured interactions, MI identified non-linear relationships, and SFI evaluated the individual predictive strength of each feature.

\paragraph{Variance Inflation Factor (VIF)}
We quantified multicollinearity using the Variance Inflation Factor (VIF), defined as:

\begin{equation}
    VIF_j = \frac{1}{1 - R_j^2}
\end{equation}

where \( R_j^2 \) is the coefficient of determination when regressing feature \( j \) on all other features. The generally accepted thresholds for VIF are:

\begin{itemize}
    \item \(VIF = 1\): No multicollinearity.
    \item \(1 < VIF < 5\): Moderate multicollinearity, generally acceptable.
    \item \(VIF > 5\): High multicollinearity, potentially problematic.
\end{itemize}

We set a threshold of 10 to flag features for removal, balancing the need to address multicollinearity with maintaining a sufficient number of features for model stability and interpretability.

\paragraph{Iterative Removal of Features}
We refined the feature set by iteratively removing features with high VIF and low combined importance:
\begin{itemize}
\item We identified features with VIF values exceeding the threshold.
\item We removed the feature with the lowest combined importance score among those flagged.
\end{itemize}

\begin{algorithm}[H]
    \caption{Iterative Feature Selection}
    \label{algorithm: iterative_feature_selection}
    \begin{algorithmic}
        \STATE \textbf{Input:} \textit{features}, \textit{target}
        \STATE \textbf{Output:} \textit{features}
        \STATE $vif_{th} \gets 10$
        \STATE $corr_{th} \gets 0.8$
        
        \STATE \textbf{Step 1: Correlation Filtering}
        \STATE $features \gets \text{CORR\_FILTER}(features, target, corr_{th})$
        
        \REPEAT
            \STATE \textbf{Step 2: Calculate Importance Metrics}
            \STATE $pfi \gets \text{PERM\_IMP}(features, target)$
            \STATE $mi \gets \text{MUT\_INFO}(features, target)$
            \STATE $sfi \gets \text{SINGLE\_FEAT\_IMPACT}(features, target)$
            
            \STATE \textbf{Step 3: Compute VIF}
            \STATE $vif \gets \text{VIF}(features)$
            
            \STATE \textbf{Step 4: Combine Importances}
            \STATE $comb\_imp \gets \text{COMB\_IMP}(pfi, mi, sfi)$

            \STATE \textbf{Termination Condition:}
            \IF{$\max(vif) \leq vif_{th}$}
                \STATE \textbf{Return:} \textit{features}
            \ELSE
                \STATE \textbf{Step 5: Identify and Remove Features}
                \STATE $high\_vif \gets \{f \mid vif[f] > vif_{th}\}$
                \STATE $imp_{min} \gets \min(\{comb\_imp[f] \mid f \in high\_vif\})$
                
                \FOR{each $f \in high\_vif$}
                    \IF{$comb\_imp[f] = imp_{min}$}
                        \STATE $features \gets features \setminus \{f\}$ \COMMENT{Remove feature with high VIF and lowest importance}
                        \STATE \textbf{break}
                    \ENDIF
                \ENDFOR
                
                \STATE \textbf{Step 6: Regularization for Further Refinement}
                \STATE $features \gets \text{REFINE\_WITH\_REG}(features)$
                \COMMENT{Drop features with zero coefficients}
            \ENDIF
        \UNTIL{$\max(vif) \leq vif_{th}$}
    \end{algorithmic}
\end{algorithm}

\paragraph{Regularization for Further Refinement}
We further refined the feature set by eliminating features with zero coefficients.

\paragraph{Termination Condition}
The iterative process concluded when all remaining features had VIF values below the threshold, ensuring a relevant feature set with minimal redundancy.

\subsubsection{Normalization and Scaling}
We standardized the shortlisted features to have a mean of zero and a standard deviation of one. This normalization ensured equal contribution of all features to model performance, facilitating effective comparison and analysis.

\subsection{Custom Preprocessing Pipeline}

We designed the \emph{custom preprocessing pipeline} (see Figure \ref{subfigure: custom pipeline}) to enhance data quality and model performance through several tailored preprocessing steps. This section details the unique procedures we applied, the rationale behind them, and their impact on the data and model performance.

\subsubsection{Grouping Columns}
We grouped columns into three categories to apply specialized preprocessing methods:

\begin{itemize}
\item \textbf{New Columns}: We identified columns such as $new\_cases$ and $new\_deaths$  as having weekly update patterns, which required specific handling to avoid bias.

\item \textbf{Independent Columns}: We classified columns such as $new\_tests$, $new\_vaccinations$, $reproduction\_rate$, $people\_vaccinated$, $people\_fully\_vaccinated$, $total\_boosters$, and $stringency\_index$ as not exhibiting weekly patterns and we did not consider any computational dependencies for them.

\item \textbf{Remaining Columns}: We designated the remaining columns as either constant or have computational dependencies.
\end{itemize}

\subsubsection{Preprocessing `New Columns'} 
\begin{figure}[!t]
\centering
{\includegraphics[width=\linewidth]{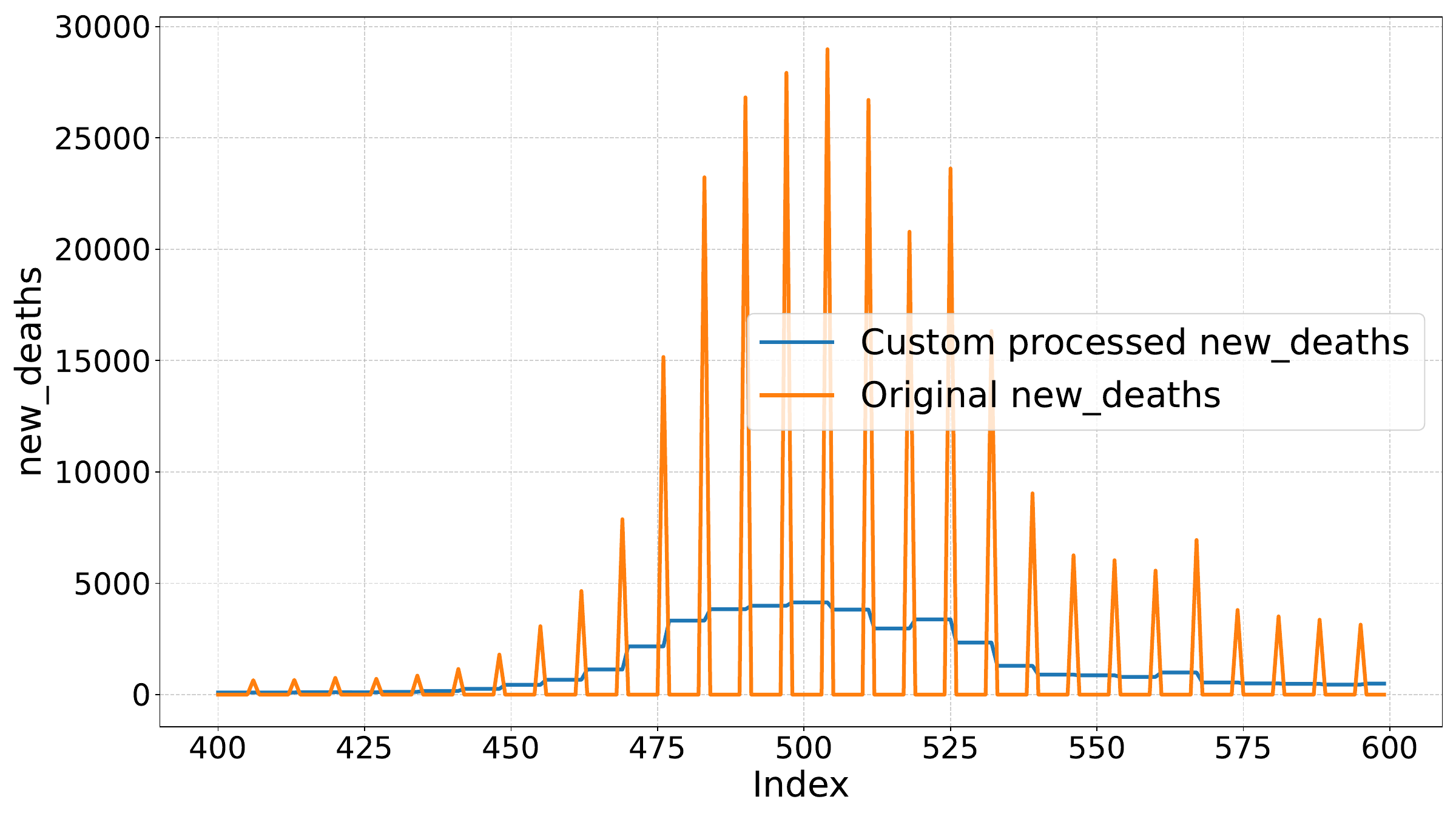}}
\caption{Comparison of Original vs. Custom Processed $new\_deaths$ Data (Zoomed-In View of Samples from index 400 to 600): Highlighting the Effectiveness of Weekly Pattern Imputation.}
\label{figure: weekly_pattern_imputation}
\end{figure}

For `New Columns', we applied the following preprocessing methods:

\paragraph{Weekly Pattern Imputation}
We corrected the weekly update pattern by redistributing total weekly values evenly across all days. This adjustment aimed to prevent bias, particularly in the $new\_deaths$ data, which showed weekly variations that could distort model predictions. Figure \ref{figure: weekly_pattern_imputation} visually represents the effectiveness of this imputation compared to the standard approach, highlighting the removal of bias introduced by weekly patterns.

\paragraph{Missing Value Imputation}
We handled missing values with linear interpolation followed by zero imputation, consistent with the standard pipeline approach to ensure comparability.

\begin{figure*}[!t]
\centering
\captionsetup[subfloat]{labelfont=scriptsize,textfont=scriptsize}
\subfloat[Global outlier detection and processing \label{subfigure: global outlier processing}]{\includegraphics[width=0.48\linewidth]{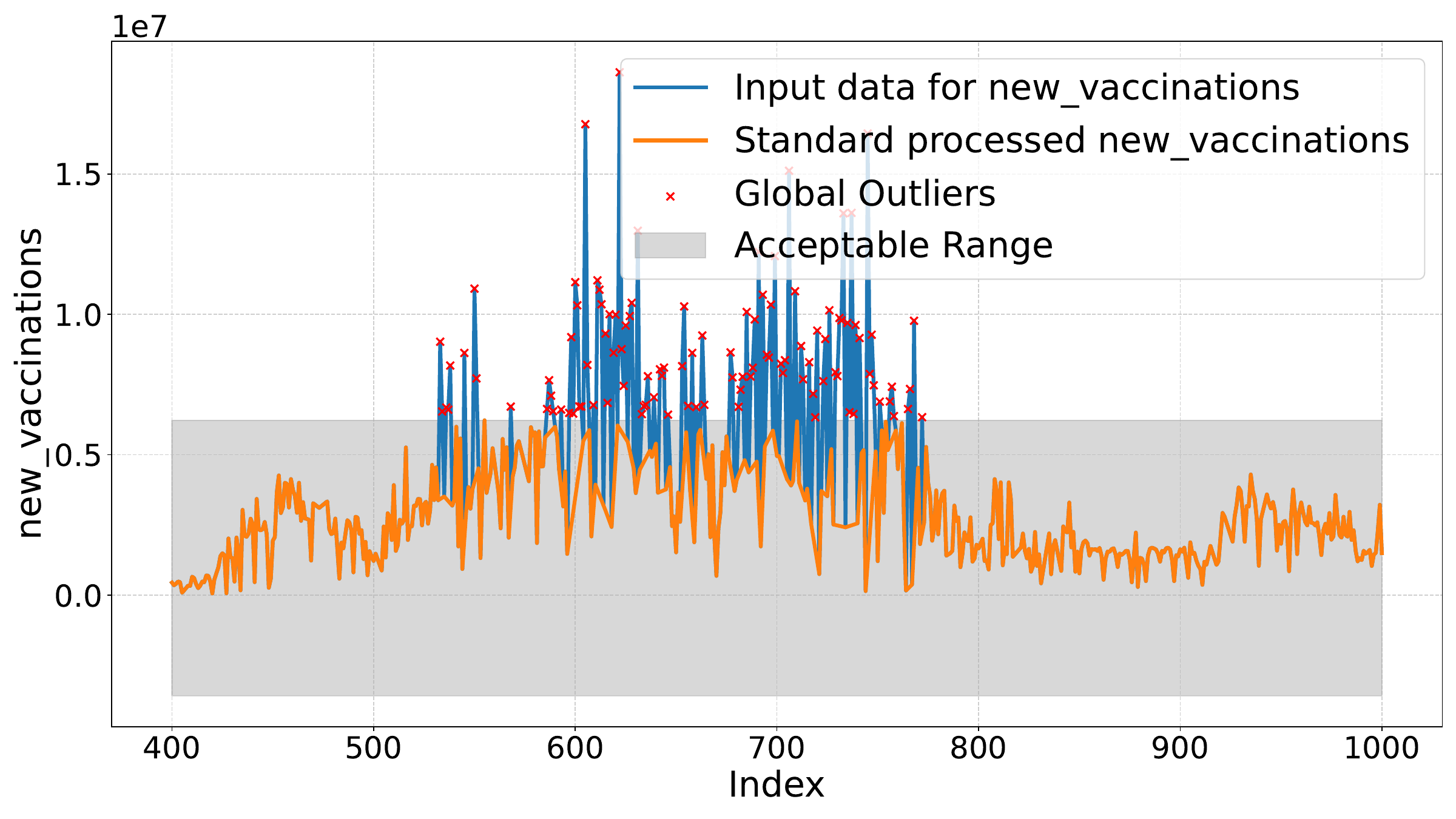}}%
\hfil
\subfloat[Local outlier detection and processing \label{subfigure: local outlier processing}]{\includegraphics[width=0.48\linewidth]{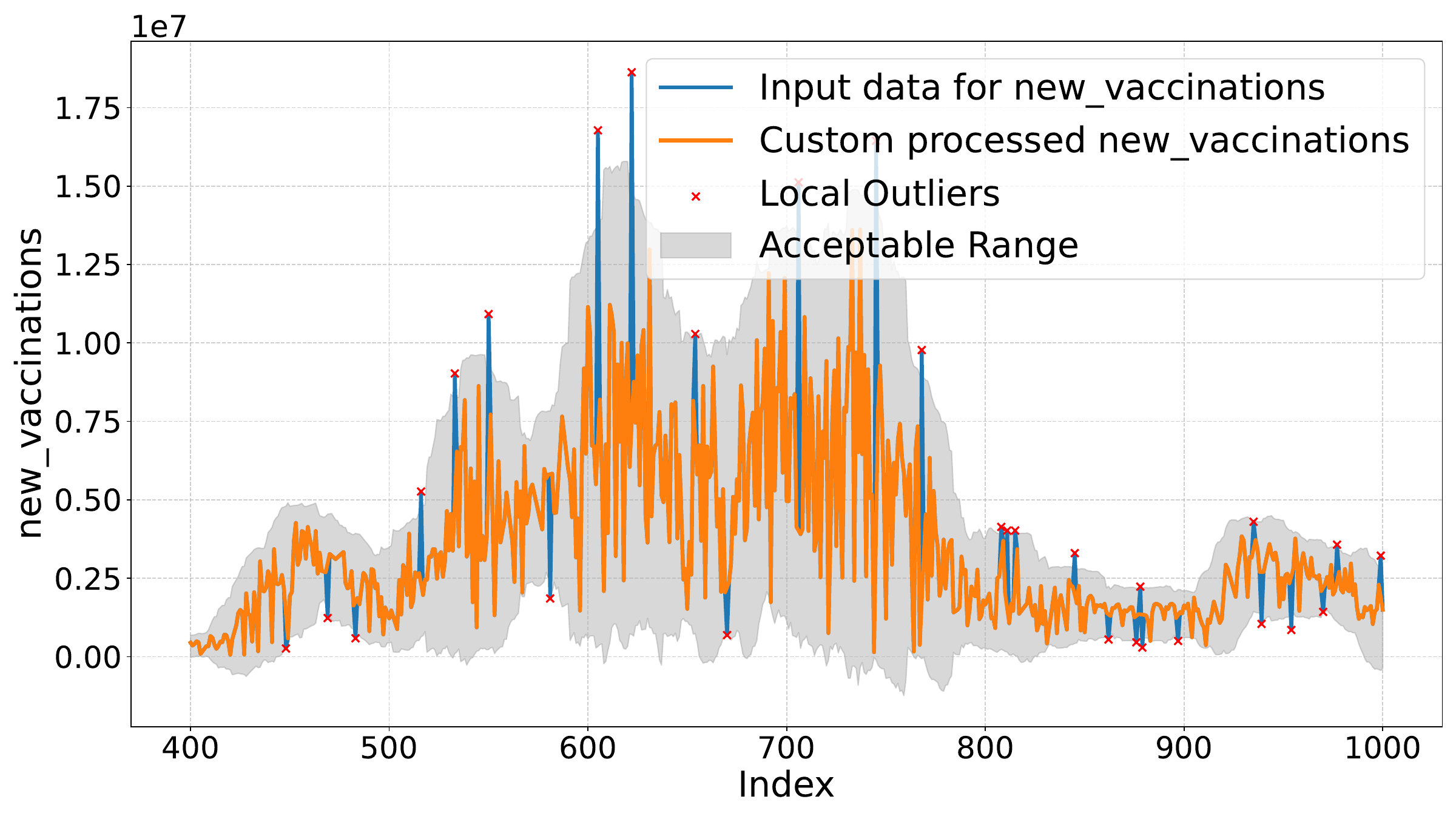}}%
\caption{Comparison of outlier detection and winsorization techniques for the $new\_vaccinations$ column from index 400 to 1000. (a) Global outlier detection and processing with the standard pipeline, and (b) Local outlier detection and processing with the custom pipeline.}
\label{figure: global vs local outlier processing}
\end{figure*}

\paragraph{Local Outlier Processing}
We addressed local extrema in time-series data by applying rolling z-scores with a 30-day window and a z-score threshold of 2. This approach effectively differentiated between genuine outliers and natural data variations, preserving critical data patterns. Figures \ref{figure: global vs local outlier processing} compare global and local outlier detection methods, demonstrating how the local approach preserved data variations and accurately detected outliers.

\subsubsection{Preprocessing `Independent Columns'} 
We applied the same preprocessing steps to `Independent Columns' as those used for `New Columns', except for weekly pattern imputation. This uniform treatment ensured consistency across different column types.

\subsubsection{Preprocessing `Remaining Columns'}
For the `Remaining Columns', we applied the following computation processing:

\paragraph{Computation Processing}

\begin{figure}[!t]
\centering
{\includegraphics[width=\linewidth]{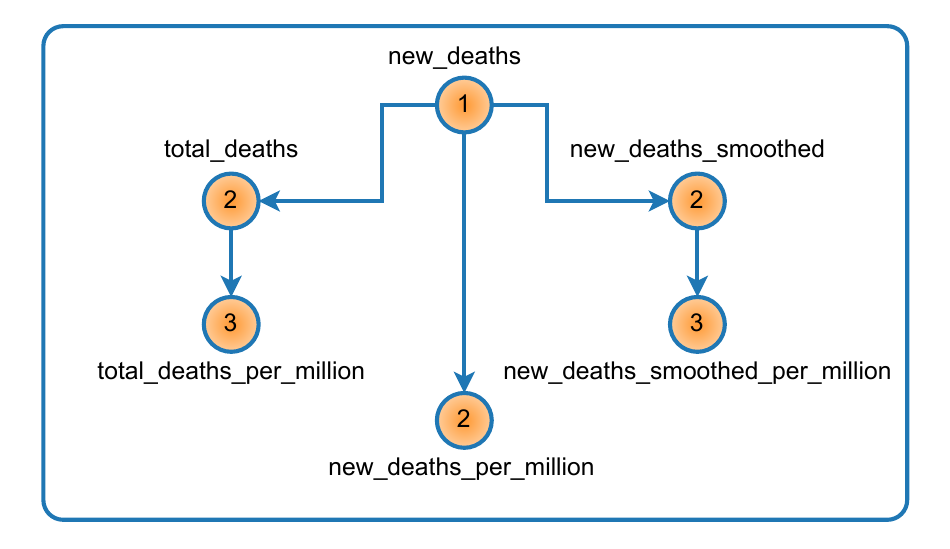}}
\caption{Column dependency graph for the `death' columns with processing orders}
\label{figure: column_dependency}
\end{figure}

We leveraged computational dependencies among columns to ensure consistency and resolve discrepancies. Figure \ref{figure: column_dependency} illustrates the dependencies and computation orders for the death-related columns. We processed $new\_deaths$ first, followed by $total\_deaths$ (using equation \ref{equation: total from new}), $new\_deaths\_smoothed$ (using equation \ref{equation: smoothed from raw}), and $new\_deaths\_per\_million$ (using equation \ref{equation: per_million from raw}) as they share the same processing order (order 2). Subsequently, we processed $total\_deaths\_per\_million$ and $new\_deaths\_smoothed\_per\_million$ (using equation \ref{equation: per_million from raw}) with processing order 3. Adhering to these manually determined but logically sequenced processing orders ensured both the consistency of computations and the integrity of the data.

\begin{figure}%
    \centering
    {\includegraphics[width=\linewidth]{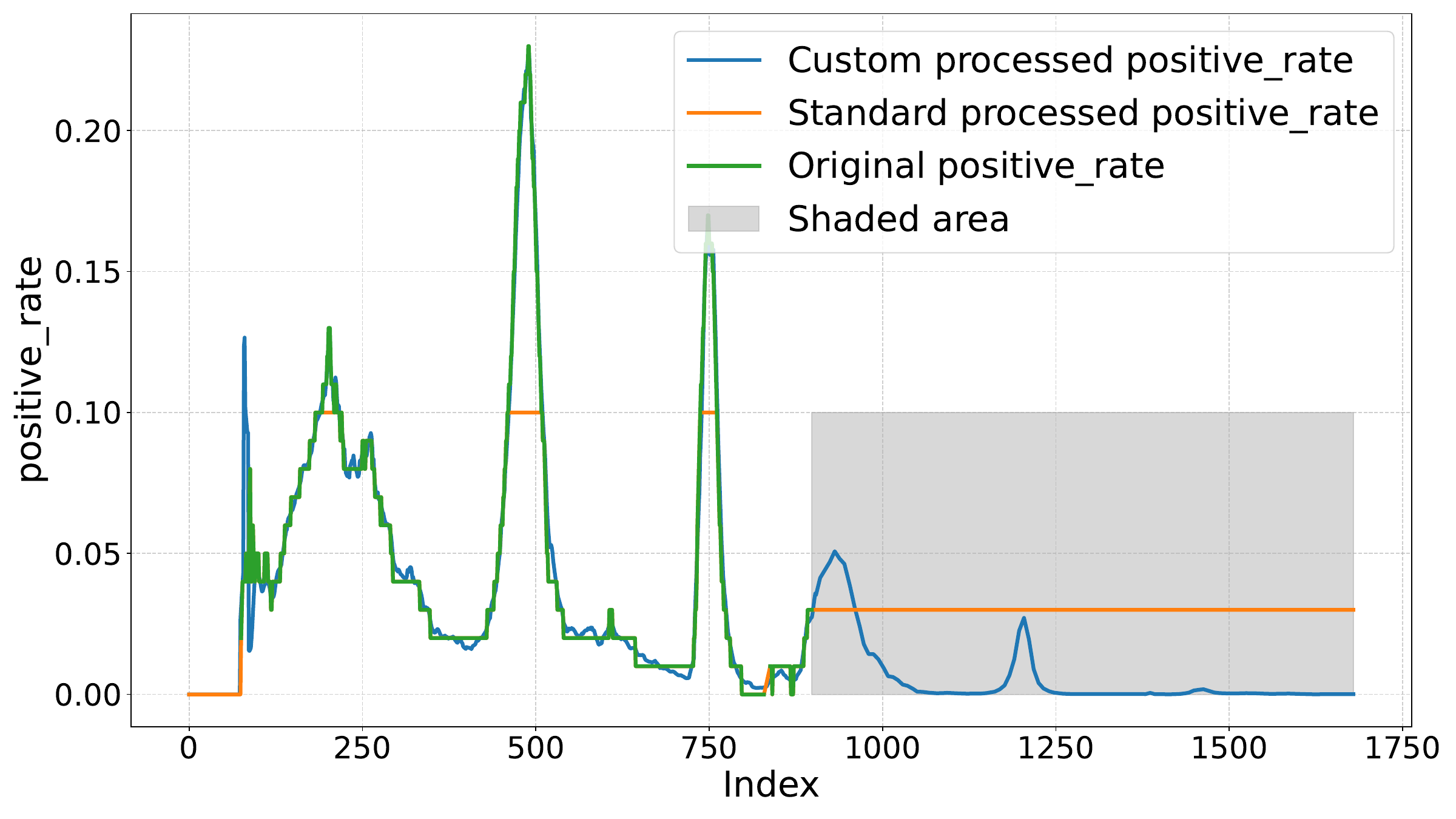}}
    \caption{Plotting for computation processing of $positive\_rate$ column}
    \label{figure: computation processing of positive_rate}
    \end{figure}

Figure \ref{figure: computation processing of positive_rate} compares the standard imputation method and our computation-based approach for the $positive\_rate$ column. While the standard pipeline imputed missing values with constant values (depicted by the orange line), our custom method calculated $positive\_rate$ using equation \ref{equation: positive_rate from new_cases and new_tests}. This approach, which relies on processed $new\_cases$ and $new\_tests$ values, delivered accurate and consistent results that captured natural variations (particularly within the shaded region) rather than relying on constant extrapolation.

\paragraph{$new$ Columns}
Computed as the difference between $total$ column values at the current index $t$ and the previous index $t-1$:
\begin{equation}
    X_{new}^{t} =
    \begin{cases}
        X_{total}^{t} & {\text{, if }} t = 0 \\
        X_{total}^{t} - X_{total}^{t-1} & {\text{, if }} t > 0
    \end{cases}
    \label{equation: new from total}
\end{equation}
We created a new column $new\_people\_vaccinated$ in our custom approach and used this equation to compute $new\_people\_vaccinated$ from $people\_vaccinated$.

\paragraph{$total$ Columns}
Computed as the cumulative sum of the corresponding $new$ column:
\begin{equation}
    X_{total}^{t} = \sum_{i=0}^{t} X_{new}^{i}
    \label{equation: total from new}
\end{equation}
This equation calculated $total$ columns from their corresponding $new$ columns, such as $new\_cases$, $new\_deaths$, $new\_tests$, and $new\_vaccinations$.

\paragraph{$positive\_rate$}
Represented the proportion of positive test results among the total tests conducted, calculated from $new\_cases$ and $new\_tests$:
\begin{equation}                
    X_{positive\_rate}^{t} = \frac{\sum_{i=0}^{6} \frac{X_{new\_cases}^{t-i}}{X_{new\_tests}^{t-i}}}{7}
    \label{equation: positive_rate from new_cases and new_tests}
\end{equation}
Refer to Figure \ref{figure: computation processing of positive_rate} for a visualization of the $positive\_rate$ computation.

\paragraph{$tests\_per\_case$}
Represented the number of tests conducted per confirmed case of COVID-19, calculated as the reciprocal of $positive\_rate$:
\begin{equation}
    X_{tests\_per\_case}^{t} = \frac{1}{X_{positive\_rate}^{t}}
    \label{equation: tests_per_case from positive_rate}
\end{equation}

\paragraph{$smoothed$ Column}
Calculated as the 7-day moving average of the corresponding $raw$ column:
\begin{equation}
    X_{smoothed}^{t} = \frac{\sum_{i=0}^{6} X_{raw}^{t-i}}{7}
    \label{equation: smoothed from raw}
\end{equation}
This equation is applied to columns such as $new\_cases$, $new\_deaths$, $new\_tests$, $new\_vaccinations$, and $new\_people\_vaccinated$.

\paragraph{$per\_million$ Column}
Calculated by dividing each $raw$ value by the total population and multiplying by one million:
\begin{equation}
    X_{per\_million}^{t} = \frac{X_{raw}^{t} \times 10^6}{population}
    \label{equation: per_million from raw}
\end{equation}
This equation was used to derive the $per\_million$ values for columns such as $total\_cases$, $new\_cases$, $total\_deaths$, $new\_deaths$, $new\_deaths\_smoothed$, $new\_cases\_smoothed$, and $new\_vaccinations\_smoothed$.

\paragraph{$per\_thousand$ Column}
Calculated by dividing each $raw$ value by the total population and multiplying by one thousand:
\begin{equation}
    X_{per\_thousand}^{t} = \frac{X_{raw}^{t} \times 10^3}{population}
    \label{equation: per_thousand from raw}
\end{equation}
This formula was used to derive $per\_thousand$ values for columns such as $total\_tests$, $new\_tests$, and $new\_tests\_smoothed$.

\paragraph{$per\_hundred$ Column}
Calculated by dividing each $raw$ value by the total population and multiplying by one hundred:
\begin{equation}
    X_{per\_hundred}^{t} = \frac{X_{raw}^{t} \times 10^2}{population}
    \label{equation: per_hundred from raw}
\end{equation}
This formula was used to derive $per\_hundred$ values for columns such as $total\_vaccinations$, $people\_vaccinated$, $people\_fully\_vaccinated$, $total\_boosters$, and $new\_people\_vaccinated\_smoothed$.

After preprocessing, we split the dataset into training, validation, and test sets. The training data then underwent iterative feature selection, followed by normalization and scaling, using the same methods as in the standard pipeline.

\subsection{Model Training and Evaluation}
We trained and evaluated ten regression models using the preprocessed datasets from both pipelines: Linear Regression, Ridge Regression, Lasso Regression, ElasticNet Regression, Support Vector Regression (SVR), Random Forest Regression, Gradient Boosting Regression, Decision Tree Regression, K-Nearest Neighbors Regression (KNN), and Neural Network Regression (Multilayer Perceptron). To prevent overfitting and optimize model performance, we applied 5-fold cross-validation and hyperparameter tuning.

\subsubsection{Evaluation Results}
We assessed the models' performance using three key metrics: $RMSE$, $R^2$, and $\textit{RMSE Variance}$.

$RMSE$ (Root Mean Squared Error) and $R^2$ (Coefficient of Determination) are standard metrics for evaluating model accuracy and explanatory power. To assess the consistency of a model's performance across different data splits, we introduced an additional metric, $\textit{RMSE Variance}$. This metric measures how consistently a model performs across the training, validation, and testing datasets. It is computed as follows:
\begin{equation}
    \label{equation: RMSE Variance}
    \textit{RMSE Variance} = \frac{1}{n} \sum_{i=1}^{n} \left( RMSE_i - \overline{RMSE} \right)^2
\end{equation}
where $RMSE_i$ represented the $RMSE$ for the training, validation, and testing datasets, $\overline{RMSE}$ was the mean $RMSE$ across these datasets, and $n$ was the total number of datasets (which is 3 in this case). $\textit{RMSE Variance}$ thus measured the variance of these three $RMSE$ values. A lower $\textit{RMSE Variance}$ indicated consistent performance across datasets, suggesting better generalizability and robustness, while a higher variance might have indicated potential overfitting or poor generalization.

By employing these metrics, we gained insights into the overall accuracy, generalization capability, and performance consistency of the models across different stages. The evaluations were conducted separately for the standard and custom preprocessing pipelines, enabling a direct comparison of the impact each pipeline had on model performance.

\subsection{Implementation Details}
The implementation was carried out in Python using Jupyter Notebooks, executed on Google Colab in a CPU environment. The complete notebook runs in approximately one hour on a standard Google Colab CPU instance (Intel Xeon CPU @ 2.20GHz with 12.6 GB of RAM).

\subsubsection{Source Code}
The complete source code for this study is available in a Jupyter Notebook and can be accessed through the \href{https://github.com/dassangita844/Preprocessing_COVID-19_Dataset_India}{Github repository}.

\section{Results and Discussion}
\label{section: Results And Discussion}

\subsection{Model Performance Evaluation}

This section presents and analyzes the performance metrics of various regression models evaluated using standard and custom preprocessing pipelines.

\subsubsection{Test Results: Test RMSE and R²}

Table \ref{table: model_performance_evaluation} summarizes the evaluation results for each model in both pipelines (for further details, see the \href{https://github.com/dassangita844/Preprocessing_COVID-19_Dataset_India}{GitHub repository}). The custom preprocessing pipeline consistently outperformed the standard one across all models. The \textit{MLPRegressor} with the custom pipeline achieved the best performance, recording an RMSE of 66.556 and an R² of 0.991. In contrast, the \textit{DecisionTreeRegressor} in the standard pipeline had the best results among the standard models, with an RMSE of 222.858 and an R² of 0.817. The custom pipeline’s lower RMSE and higher R² values indicate superior predictive accuracy and a better model fit across the board.

\begin{table}[!t]
\centering
\caption{Performance Metrics for Various Models Using Standard and Custom Pipelines}
\resizebox{\linewidth}{!}{
\begin{tabular}{@{}llrrr@{}}
\toprule
Pipeline & Model & Test RMSE & Test R²   & RMSE Variance \\

\midrule
\multirow{10}{*}{\centering Standard} 
& \textbf{DecisionTreeRegressor} & \textbf{222.858} & \textbf{0.817} & 776.666 \\
& RandomForestRegressor & 238.373 & 0.790 & 3740.121 \\
& GradientBoostingRegressor & 242.053 & 0.784 & 1762.070 \\
& SVR & 278.971 & 0.713 & 746.453 \\
& KNeighborsRegressor & 366.496 & 0.504 & 8318.961 \\
& Ridge & 406.051 & 0.391 & 1437.239 \\
& ElasticNet & 406.054 & 0.391 & 1437.262 \\
& LinearRegression & 406.144 & 0.391 & 1438.698 \\
& Lasso & 406.185 & 0.391 & 1441.244 \\
& MLPRegressor & 419.340 & 0.350 & \textbf{13739.921} \\

\midrule
\multirow{10}{*}{\centering Custom} 
& \textbf{MLPRegressor} & \textbf{66.556} & \textbf{0.991} & \textbf{52.125} \\
& KNeighborsRegressor & 84.510 & 0.985 & 210.551 \\
& GradientBoostingRegressor & 86.926 & 0.984 & 108.862 \\
& RandomForestRegressor & 144.046 & 0.956 & 2126.466 \\
& DecisionTreeRegressor & 146.775 & 0.955 & 1768.430 \\
& SVR & 208.737 & 0.908 & 766.496 \\
& Ridge & 406.188 & 0.652 & 6.828 \\
& LinearRegression & 406.204 & 0.652 & 6.889 \\
& Lasso & 406.322 & 0.652 & 6.967 \\
& ElasticNet & 406.342 & 0.652 & \textbf{6.772} \\

\bottomrule
\end{tabular}
}
\label{table: model_performance_evaluation}
\end{table}

\subsubsection{Generalizability: Overfitting and Underfitting}

The consistency of model performance across different datasets was assessed using RMSE variance, where lower values indicate better generalization. Models from the custom pipeline generally exhibited lower RMSE variances, suggesting they were less prone to overfitting. For instance, the \textit{MLPRegressor} in the custom pipeline demonstrated high stability with an RMSE variance of 52.125, while the same model in the standard pipeline showed significant instability with an RMSE variance of 13,739.921, indicating potential overfitting.

\subsubsection{Impact of Weekly Pattern Imputation}

The dataset initially displayed a weekly update pattern for $new\_deaths$, with zeros reported for six days and the total on the seventh day, This pattern could bias models by distorting the underlying trend. The custom pipeline addressed this by redistributing the weekly totals across all days, which enhanced model performance. Although we tested $new\_deaths\_smoothed$ as a target variable, results indicated that the standard pipeline performed unusually well with this target due to global outlier processing. This processing stripped away essential data variations, inflating performance metrics artificially, as shown in Figure \ref{subfigure: global outlier processing}. This finding underscores the importance of preserving data variability for accurate model evaluation.

\subsubsection{Global vs. Local Outlier Processing}

Global outlier detection and processing, using fixed z-score thresholds, often fails to capture local data variability. Our findings suggest that local outlier detection, which adapts thresholds contextually, better preserves data integrity and improves model accuracy. This approach is preferred for time-series data like COVID-19.

\subsubsection{Computation Processing and Feature Stability}

Computation processing ensured consistent relationships between features, leading to more stable models. Early iterations of the feature selection exhibited infinite Variance Inflation Factor (VIF) values for most columns (see the \href{https://github.com/dassangita844/Preprocessing_COVID-19_Dataset_India}{GitHub repository}), indicating perfectly consistent feature relationships. In contrast, the standard pipeline showed finite VIF values, reflecting less consistent feature dependencies and contributing to less reliable model performance.

\subsubsection{Iterative Feature Selection and its Impact}

Iterative feature selection was employed to refine the feature set, reducing the number of numeric features from 34 (excluding the target) to 5 in the custom pipeline and 7 in the standard pipeline. Despite this reduction, the custom pipeline achieved higher accuracy and consistency, demonstrating the effectiveness of the feature selection approach. The iterative process effectively managed feature relationships and minimized multicollinearity, as indicated by VIF values below 5 (see Table \ref{table:shortlisted_features}) in both pipelines. However, the custom pipeline's features exhibited superior combined importance scores, contributing to its enhanced performance.

\paragraph{Feature Importance Scores}
Table \ref{table:shortlisted_features} compares the combined feature importance scores and VIF values for features selected by the standard and custom pipelines. This table highlights the differences in feature importance and multicollinearity between the pipelines. Features in the custom pipeline have significantly higher combined importance scores compared to those in the standard pipeline, reflecting more effective data processing. For instance, the $stringency\_index$ was shortlisted by both pipelines, but its combined importance score was 0.160 in the standard pipeline compared to 1.128 in the custom pipeline. This discrepancy underscores the substantial performance improvement achieved with the custom pipeline.

\begin{table}[!t]
\centering
\caption{Combined Feature Importance Scores and VIF Values for Shortlisted Features in Standard and Custom Pipelines}
\resizebox{0.95\linewidth}{!}{
\begin{tabular}{@{}llcc@{}}
\toprule
Pipeline & Features & Combined Importance & VIF \\ 

\midrule
\multirow{6}{*}{\parbox{0.1\linewidth}{\centering Standard}} 
& $new\_cases\_per\_million$ & 0.807 & 1.096 \\
& $new\_deaths\_smoothed$ & 0.353 & 3.197 \\
& $total\_cases\_per\_million$ & 0.328 & 3.013 \\
& $tests\_per\_case$ & 0.242 & 3.131 \\
& $new\_people\_vaccinated\_smoothed$ & 0.237 & 4.718 \\
& $new\_vaccinations$ & 0.190 & 4.998 \\
& $stringency\_index$ & 0.160 & 4.852 \\
\midrule
\multirow{5}{*}{\parbox{0.1\linewidth}{\centering Custom}} 
& $new\_cases$ & 2.161 & 1.570 \\
& $total\_deaths\_per\_million$ & 1.676 & 2.464 \\
& $stringency\_index$ & 1.128 & 2.696 \\
& $tests\_per\_case$ & 1.001 & 1.395 \\
& $new\_people\_vaccinated$ & 0.708 & 1.464 \\
\bottomrule
\end{tabular}
}
\label{table:shortlisted_features}
\end{table}

The custom pipeline’s features exhibited infinite VIF values, indicating stable relationships, whereas the standard pipeline’s finite VIF values suggested less effective feature selection.

\subsection{Summary and Recommendations}

\paragraph{Weekly Pattern Imputation}
Redistributing weekly totals to daily updates improved model performance by preserving data integrity and minimizing bias. We recommend this approach for accurate and reliable predictions.

\paragraph{Local Outlier Processing}
Local outlier processing effectively identifies and handles outliers while preserving data variance. This method is preferred over global outlier processing for consistent and accurate modelling.

\paragraph{Computation Processing}
The custom pipeline's computation processing ensured perfect feature consistency, leading to superior model performance. Identifying and leveraging computational dependencies is essential for accurate imputation and modelling.

\paragraph{Iterative Feature Selection}
The custom pipeline’s iterative feature selection, despite using fewer features, achieved higher accuracy and consistency due to the effective management of feature relationships and multicollinearity. This approach is recommended for optimizing feature sets in predictive modelling.

In conclusion, the custom pipeline's tailored approaches in imputation, outlier processing, computation, and feature selection resulted in more robust and accurate predictive models. These methods not only demonstrated superiority over the standard pipeline but also have broad applicability to other preprocessing and machine learning tasks. Adopting these custom approaches can enhance predictive modelling across various domains and datasets.

\section{Conclusion}
\label{section: Conclusion}

This study highlights the essential role of a tailored data preprocessing pipeline in enhancing the accuracy and reliability of COVID-19 mortality predictions. By redistributing weekly totals for the target variable ($new\_deaths$), utilizing local outlier detection to preserve data variance, leveraging computational dependencies, and employing iterative feature selection, the custom pipeline significantly outperformed the standard approach. Specifically, the \textit{MLPRegressor} model with the custom pipeline achieved an RMSE of 66.556, an R² of 0.991, and an RMSE variance of 52.125, compared to the best standard pipeline results with the \textit{DecisionTreeRegressor}, which recorded an RMSE of 222.858, an R² of 0.817, and an RMSE variance of 776.666, highlighting the substantial improvement. These improvements demonstrate the effectiveness of custom preprocessing, particularly for time-series data with distinct patterns and outliers. Although this study focuses on data from India, future research should apply this pipeline to datasets from other countries, where additional dependencies may need to be addressed. The techniques of local outlier detection, iterative feature selection, and RMSE variance evaluation are broadly applicable and can enhance data integrity, consistency, and model performance across diverse domains and datasets.
%
\section*{}
\bibliographystyle{ieeetr}
\bibliography{bibliography}
\vfill



\newpage

\end{document}